\documentclass{article}

\DeclareUnicodeCharacter{00A0}{ }

\usepackage[english]{babel}

\usepackage[letterpaper,top=2cm,bottom=2cm,left=3cm,right=3cm,marginparwidth=1.75cm]{geometry}

\usepackage{authblk}

\usepackage{amsmath}
\usepackage{graphicx}
\usepackage[colorlinks=true, allcolors=blue]{hyperref}

\usepackage{tabularx}
\usepackage{amssymb,amsfonts} 
\usepackage{algorithmic}      
\usepackage{textcomp}
\usepackage{xcolor}
\usepackage{bbm}              
\usepackage{booktabs}         
\usepackage{subcaption}       
\usepackage{float}            
\usepackage{caption}          

\title{Shifting Uncertainty to Critical Moments: Towards Reliable Uncertainty Quantification for VLA Model}

\author{
  Yanchuan Tang$^{1*\dagger}$,
  Taowen Wang$^{1}$\thanks{These authors contributed equally to this work. \newline  \phantom{$^{**}$} Work done during Yanchuan Tang's and Taowen Wang's remote internship at Rutgers University.},
  Yuefei Chen$^{1}$,
  Boxuan Zhang$^{1}$,
  Qiang Guan$^{2}$\\
  \vspace{-10pt}
  {\large Ruixiang Tang$^{1}$\thanks{Corresponding author: ruixiang.tang@rutgers.edu}}\\
  {\large$^{1}$Rutgers University \quad $^{2}$Kent State University}
}

\begin{document}
\maketitle

\begin{abstract}
Vision-Language-Action (VLA) models enable general-purpose robotic policies by mapping visual observations and language instructions to low-level actions, but they often lack reliable introspection. A common practice is to compute a token-level uncertainty signal and take its mean over a rollout. However, mean aggregation can dilute short-lived but safety-critical uncertainty spikes in continuous control. In particular, successful rollouts may contain localized high-entropy segments due to benign noise or non-critical micro-adjustments, while failure rollouts can appear low-entropy for most timesteps and only exhibit brief spikes near the onset of failure. We propose a unified uncertainty quantification approach for predicting rollout success versus failure that (1) uses max-based sliding window pooling to preserve transient risk signals, (2) applies motion-aware stability weighting to emphasize high-frequency action oscillations associated with unstable behaviors, and (3) performs DoF-adaptive calibration via Bayesian Optimization to prioritize kinematically critical axes. Experiments on the LIBERO benchmark show that our method substantially improves failure prediction accuracy and yields more reliable signals for failure detection, which can support downstream human-in-the-loop interventions.
\end{abstract}


\section{Introduction}
\label{sec:intro}

Vision-Language-Action (VLA) models have emerged as a promising direction for developing general-purpose robotic policies, using large language models as backbones to map visual observations and language instructions directly to low-level control signals~\cite{zitkovich2023rt, team2024octo, kim2024openvla, black2024pi_0, black2025pi_}. Despite their impressive scalability, these models often lack built-in mechanisms for introspection. Errors in robotic decision-making can lead to physical collisions, safety hazards, or hardware damage. Therefore, Uncertainty Quantification (UQ), which enables the model to estimate its confidence and identify potential failures before they occur, is essential for safe deployment and reliable human-robot collaboration.

Few pioneering studies have attempted to understand uncertainty in VLA models. For example, $\pi_0$-FAST leverages token-level entropy to trigger human intervention when uncertainty is detected~\cite{karli2025ask}. While insightful, these efforts mainly apply existing LLM uncertainty metrics, such as entropy averaged over tokens. Our empirical analysis suggests that this direct application of LLM-style metrics to the continuous domain of robotics faces a significant challenge, which we term the \textbf{Averaging Trap}. In robotic manipulation, successful trajectories often contain high-entropy segments due to benign noise or non-critical micro-adjustments. Conversely, failure cases may exhibit low entropy for the majority of execution, masking the transient spikes that actually cause the failure. As a result, global statistics or simple thresholds often fail to distinguish between benign fluctuations and fatal errors, leading to performance that can be indistinguishable from random guessing in precise spatial tasks.

Our motivation arises from the limitations of global, uniform uncertainty estimation in capturing the nuanced dynamics of robotic control. We argue that effective introspection should focus on local temporal windows to preserve transient failure signals rather than averaging uncertainty over an entire episode, a perspective supported by recent advances in related domains~\cite{song2025discovering, chen2025tskan}. In addition, physical instability, reflected in high-frequency action oscillations~\cite{li2025frequency}, serves as an important indicator of uncertainty and should be emphasized accordingly. Finally, different degrees of freedom (DoF) contribute unequally to task risk, which calls for an adaptive weighting mechanism. Our approach, therefore, consists of three key components:

\begin{itemize}
    \item \textbf{Sliding Window Pooling:} Instead of global averaging, we employ a max-based temporal sliding window to capture local uncertainty density. This allows the system to detect transient spikes hidden within long-horizon trajectories.
    \item \textbf{Motion-Aware Stability Weighting:} We design a dynamic weighting scheme, which we term Action Transfer Reweighting (ATR), based on action stability. Since uncertainty often appears as erratic motion, actions with high-frequency fluctuations receive higher uncertainty weights, while smooth and stable trajectories are down-weighted, aligning model logits with the physical behavior of the end-effector.
    \item \textbf{DoF-Adaptive Calibration with Bayesian Optimization:} Since different joints contribute unequally to task risk, we move beyond uniform uncertainty averaging. We use Bayesian Optimization to automatically learn adaptive weights for each DoF, producing a calibrated uncertainty metric that prioritizes more critical axes according to the task’s kinematic requirements.
\end{itemize}

We evaluate our framework on the widely used LIBERO~\cite{liu2023libero} benchmark. Results show that our method effectively captures model uncertainty, leading to more reliable failure detection and supporting safer analysis and monitoring of VLA behavior.

\begin{figure}[t]
  \centering
  \includegraphics[width=0.9\textwidth]{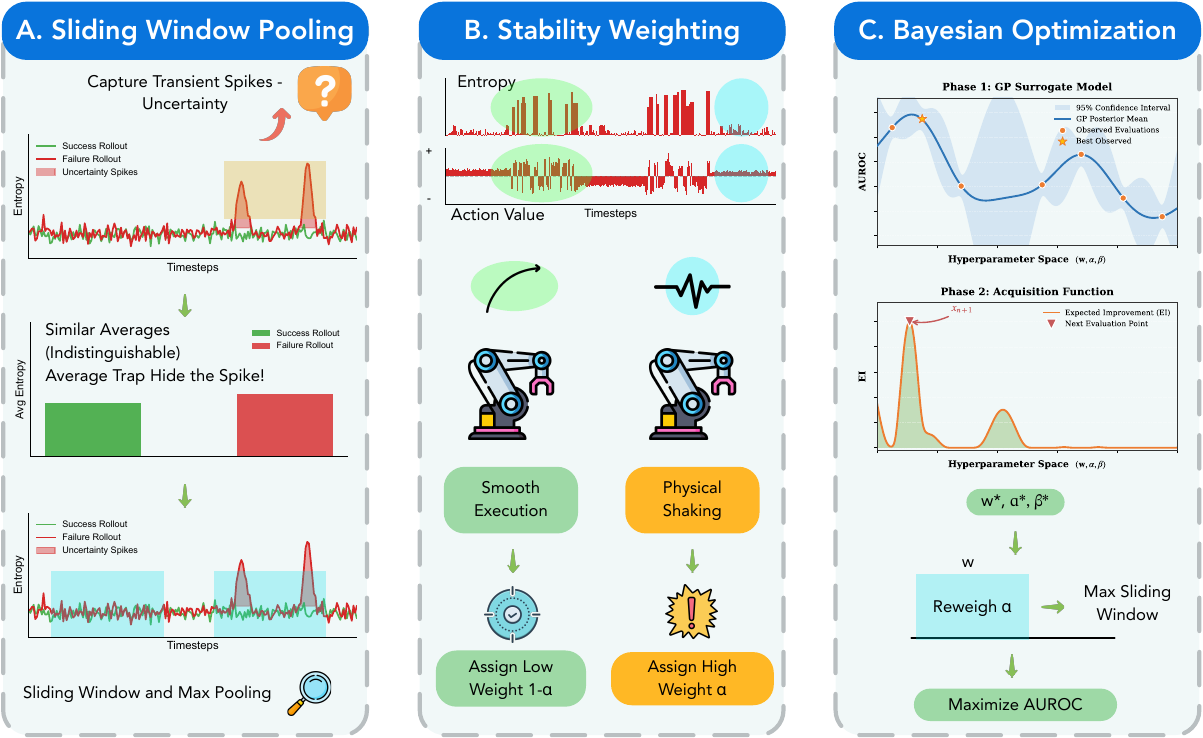}
  \caption{Overview of the proposed uncertainty quantification framework.
  Standard global averaging often masks failure signals. We propose:
  (A) Sliding Window Pooling (SW) to capture transient uncertainty spikes;
  (B) Action Transfer Reweighting (ATR) to prioritize uncertainty during oscillatory actions;
  and (C) Bayesian Optimization (BO) to learn adaptive weights for kinematically critical DoFs.}
  \label{fig:overview}
\end{figure}

\section{Related Work}
\label{sec:related}

\subsection{Vision-Language-Action Models}
Vision-Language-Action (VLA) models unify visual perception, language understanding, and action generation, enabling robots to execute manipulation tasks from high-level instructions. Recent works such as RT-2~\cite{zitkovich2023rt}, Octo~\cite{team2024octo}, and OpenVLA~\cite{kim2024openvla} demonstrate strong generalization by leveraging large-scale vision-language backbones trained on robot demonstrations. More recently, models like $\pi_0$~\cite{black2024pi_0} and $\pi_{0.5}$~\cite{black2025pi_} have advanced this paradigm by employing flow matching techniques to generate continuous, high-frequency action trajectories. Despite their success, these models are typically deployed as black boxes and lack explicit mechanisms to assess the reliability of their predicted actions, which limits their applicability in safety-critical and long-horizon settings.

\subsection{Uncertainty Metrics in Foundation Models and Robotics}
Uncertainty estimation in foundation models is commonly based on entropy- or likelihood-related measures, such as token-level entropy and perplexity, which provide signals of model confidence during autoregressive decoding~\cite{kadavath2022language, kuhn2023semantic, ling2024uncertainty, lin2023generating}. These metrics have been widely used for introspection and out-of-distribution detection in large language models~\cite{huang2025survey}. Beyond token statistics, recent work has also utilized internal representations to enhance LVLM safety~\cite{hua2025rethinking}. In robotics, uncertainty has traditionally been modeled using ensembles or Bayesian approaches~\cite{chua2018deep, hansen2023td}, but such methods are often impractical for large VLA models. Recent work has begun to extract uncertainty directly from VLA token distributions~\cite{ren2023robots, zitkovich2023rt}, typically by aggregating entropy into global statistics. However, such global aggregation can obscure localized and transient uncertainty in long-horizon control. Similarly, recent advancements in vision-language model adaptation indicate that treating all entropy signals uniformly is suboptimal, as high-entropy samples often contain critical boundary information for robust prediction calibration~\cite{chen2025multi}. Our work builds on these metrics and introduces aggregation strategies that better reflect the temporal and kinematic structure of robotic manipulation.

\section{The Averaging Trap: Why Mean-Entropy Fails for VLA Uncertainty}
\label{sec:method}

\paragraph{Task setup.}
We study rollout-level uncertainty quantification for long-horizon robotic manipulation.
A rollout is defined as a sequence $\tau=\{(\mathbf{o}_t,\mathbf{a}_t)\}_{t=1}^{T}$, where $\mathbf{o}_t$ denotes multimodal observations and $\mathbf{a}_t$ the executed low-level action at timestep $t$ over a horizon $T$. Each rollout is associated with a binary outcome $y\in\{0,1\}$, indicating task \emph{success} ($y{=}1$) or \emph{failure} ($y{=}0$).
Our objective is to compute a scalar uncertainty score $S(\tau)$ that reflects the risk of failure of an entire rollout, such that failed executions are assigned consistently higher scores than successful ones.
This score is intended to serve as a reliable trigger for intervention, enabling early termination or human-in-the-loop control before irreversible failures occur.

\paragraph{Per-step per-DoF token entropy.}
In OpenVLA, each action $\mathbf{a}_t \in \mathbb{R}^7$ corresponds to a 7-DoF Cartesian task-space command,
consisting of delta end-effector motions and a binary gripper control signal, i.e., $\mathbf{a}_t=\{a_{t,d}\}_{d=1}^{7}$.
Specifically, for each dimension $d\in\{1,\dots,7\}$, the model decodes actions into a categorical distribution $p_{t,d}(\cdot)$ over 256 discrete bins. Thus, the token set $\mathcal{V}_d$ refers precisely to the logits corresponding to these 256 bins.
We quantify the uncertainty of the $d$-th action component at time $t$ using the token entropy:
\begin{equation}
H_{t,d} \;=\; -\sum_{v\in\mathcal{V}_d} p_{t,d}(v)\log p_{t,d}(v),
\label{eq:Ht_d}
\end{equation}
where $p_{t,d}(v)$ denotes the probability assigned to token $v$ for the $d$-th action dimension.
This per-DoF entropy measures the model’s uncertainty in generating the corresponding Cartesian control command
at each timestep.

\begin{figure}[t]
  \centering
  \begin{subfigure}[b]{0.48\columnwidth}
    \centering
    \includegraphics[width=\linewidth]{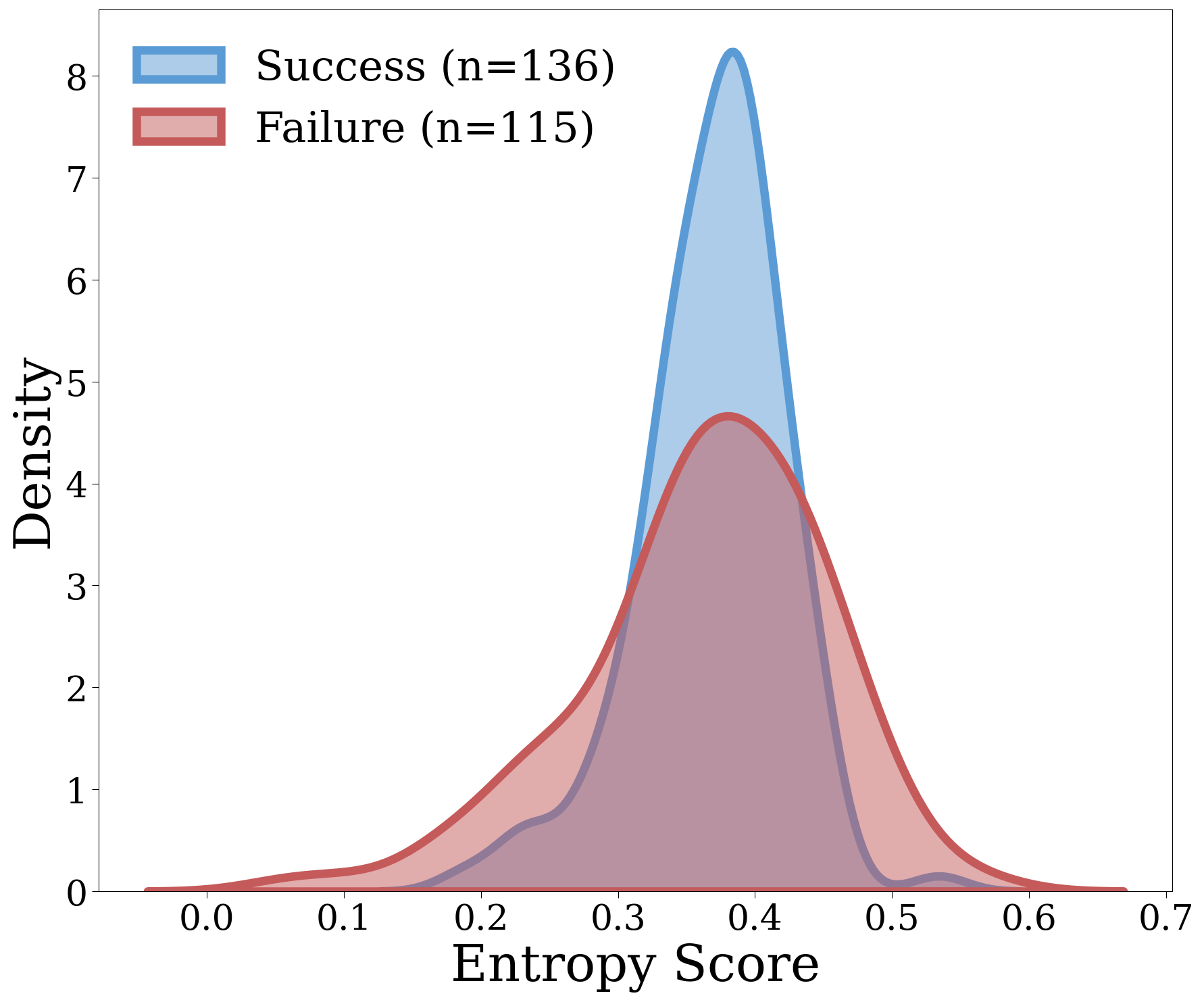}
    \caption{Train (AUROC=0.51)}
    \label{fig:trap_train}
  \end{subfigure}
  \hfill
  \begin{subfigure}[b]{0.48\columnwidth}
    \centering
    \includegraphics[width=\linewidth]{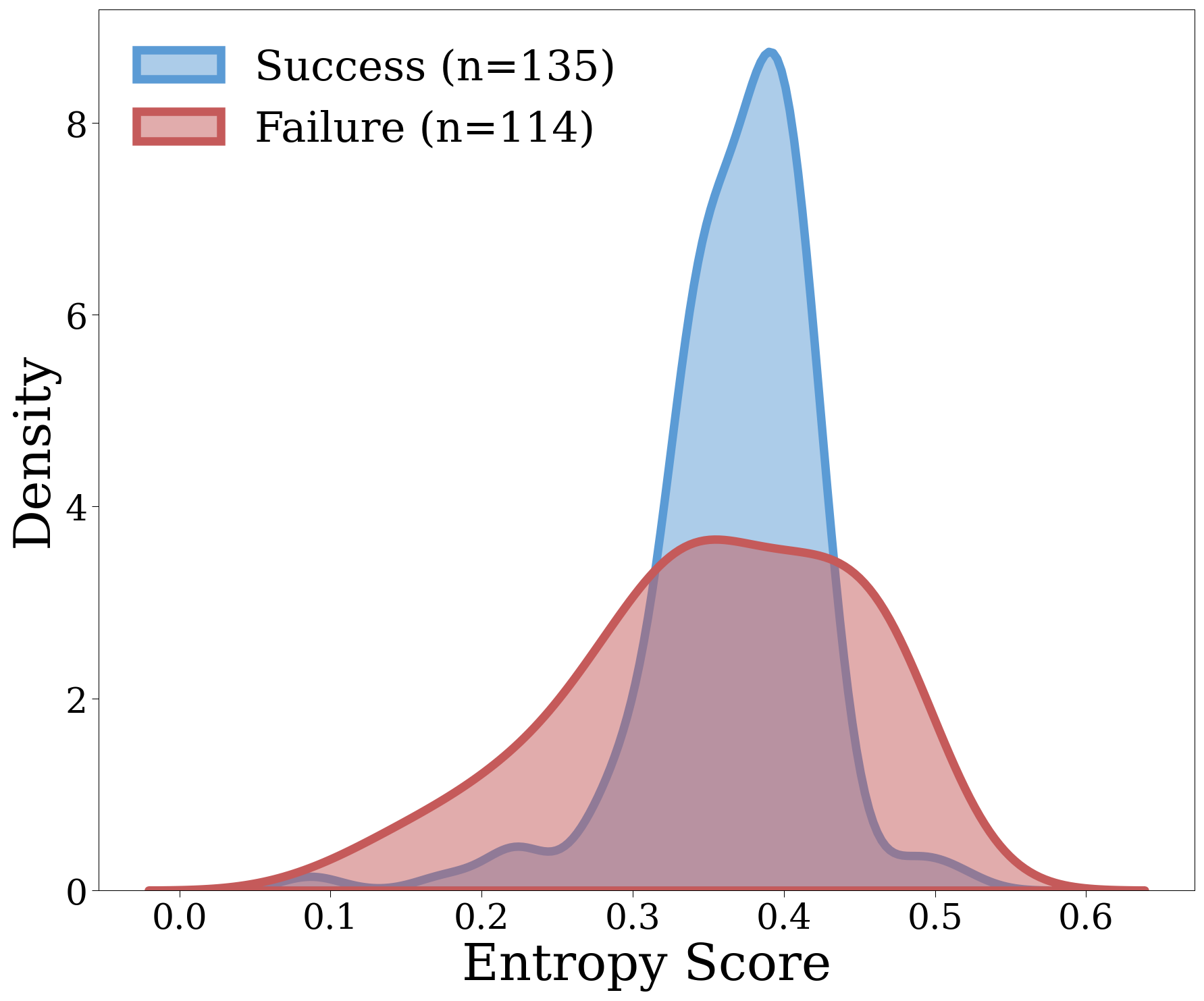}
    \caption{Test (AUROC=0.47)}
    \label{fig:trap_val}
  \end{subfigure}
  
  \caption{Empirical evidence of the Averaging Trap on LIBERO-10. The plots show the probability density of the global mean entropy $S_{\text{Avg}}(\tau)$ for success (blue) and failure (red) rollouts. The significant overlap between the two distributions, coupled with near-random AUROC scores (0.51 on train and 0.47 on test), demonstrates that global averaging masks critical failure signals and fails to distinguish between successful and failed executions.}
  \label{fig:avg_trap}
\end{figure}

\paragraph{Baseline: Global Averaging and the Averaging Trap.}
We begin with an empirical evaluation of a simple and widely used baseline that summarizes token-level uncertainty by globally averaging entropy across time and the 7 action DoFs:
\begin{equation}
S_{\text{Avg}}(\tau) \;=\; \frac{1}{7T}\sum_{t=1}^{T}\sum_{d=1}^{7} H_{t,d}.
\label{eq:avg_baseline}
\end{equation}
However, this global statistic is prone to the \emph{Averaging Trap} in continuous control.
Successful rollouts can contain localized high-entropy segments due to benign noise or non-critical micro-adjustments, which inflate $S_{\text{Avg}}$.
Conversely, many failures appear low-entropy for most timesteps and only exhibit brief spikes near the onset of failure, which are masked by the average.
As a consequence, $S_{\text{Avg}}$ often fails to reliably separate success from failure, especially in precise manipulation tasks where a short critical mistake can invalidate an entire rollout (see Fig.~\ref{fig:avg_trap}).

To escape this Averaging Trap, an effective uncertainty score should answer two questions:
\begin{itemize}
    \item \textbf{Temporal locality:} How can we aggregate uncertainty so that brief high-risk segments are preserved rather than averaged away?
    \item \textbf{Physical relevance:} How can we emphasize uncertainty that aligns with unstable action execution instead of benign fluctuations?
\end{itemize}

\section{The Proposed Method}
\label{sec:proposed_method}
Motivated by our observations, we develop an adaptive uncertainty estimation framework for VLA models. We first address \emph{temporal locality} by replacing global averaging with a sliding-window operator, and then incorporate \emph{physical relevance} via stability-aware reweighting in subsequent components.

\subsection{Sliding Window Pooling (SW)}
\label{sec:method:sw}
To preserve transient risk signals, we replace global averaging with max-based temporal pooling.
We first aggregate per-step entropy across DoF as $e_t = \frac{1}{7}\sum_{d=1}^{7} H_{t,d}$.
Given a window size $w$, we compute window means:
\begin{equation}
\bar e_t \;=\; \frac{1}{w}\sum_{k=t}^{t+w-1} e_k,\quad
t \in \{1, 2,\dots\},
\label{eq:sw_mean}
\end{equation}
and define the rollout score by max pooling over these windows: $S_{\text{SW}}(\tau) = \max_{t} \bar e_t$.
This directly targets a common failure mode in manipulation: brief local uncertainty spikes may be decisive even when the remainder of the rollout is well executed. In practice, the resulting window-level score $S_{\text{SW}}(\tau)$ can be compared against a predefined threshold
to trigger early termination or human intervention, thereby preventing irreversible failures such as object drops,
collisions, or task-invalidating errors.

\subsection{Action Transfer Reweighting (ATR)}
\label{sec:method:atr}
Token-level uncertainty becomes most informative when it coincides with physically unstable action execution.
Motivated by this observation, we introduce \textit{Action Transfer Reweighting} (ATR), which assigns higher importance to entropy values at timesteps where the executed actions exhibit oscillatory behavior, as such patterns often precede rollout failure.

We define a per-step, per-DoF instability indicator using sign inconsistency between consecutive actions:
\begin{equation}
c_{t,d} \;=\; \mathbbm{1}\!\left[\mathrm{sign}(a_{t,d}) \neq \mathrm{sign}(a_{t-1,d})\right],\quad t\ge2,
\label{eq:sign_flip}
\end{equation}
where we treat 0 as non-negative, and explicitly set $c_{1,d} = 0$ since there is no preceding action. 
Crucially, our instability indicator $c_{t,d}$ is computed directly on the VLA's \textit{inferred action commands}, rather than on proprioceptive sensor readings. Since model-predicted actions are purely computational, this binary sign-flip remains a clean and robust signal immune to downstream mechanical or sensor noise during real-world deployment. 
We then apply a stability weight:
\begin{equation}
\alpha_{t,d} \;=\;
\begin{cases}
1-\alpha, & c_{t,d}=0,\\
\alpha, & c_{t,d}=1,
\end{cases}
\label{eq:atr_weight}
\end{equation}
with hyperparameter $\alpha\in(0,1)$ controlling the contrast between stable versus oscillatory behavior. We typically choose $\alpha > 0.5$ to emphasize unstable moments.
The reweighted entropy is $\tilde H_{t,d}=\alpha_{t,d}H_{t,d}$, and the ATR rollout score is:
\begin{equation}
S_{\text{ATR}}(\tau) \;=\; \frac{1}{7T}\sum_{t=1}^{T}\sum_{d=1}^{7}\tilde H_{t,d}.
\label{eq:atr_score}
\end{equation}
Intuitively, ATR down-weights benign uncertainty occurring under stable motion, while amplifying uncertainty that coincides with shaking or hesitation patterns, 
yielding a more discriminative and physically grounded uncertainty signal.

\subsection{Unified SW+ATR Scoring}
\label{sec:method:swatr}
We combine spike-preserving pooling (SW) with motion-aware weighting (ATR) by applying max-based sliding window pooling to the reweighted entropy:
\begin{equation}
\label{eq:swatr_score}
\begin{aligned}
\tilde e_t \;&=\; \frac{1}{7}\sum_{d=1}^{7}\tilde H_{t,d}, \\
S_{\text{SW+ATR}}(\tau) \;&=\; \max_{t}\; \frac{1}{w}\sum_{k=t}^{t+w-1}\tilde e_k .
\end{aligned}
\end{equation}
SW prevents critical spikes from being averaged out, while ATR makes spikes more discriminative by anchoring them to unstable physical motion, resulting in a unified scoring function that is both temporally sensitive and physically grounded.

\subsection{DoF-Adaptive Calibration with Bayesian Optimization}
\label{sec:method:bo}
Different DoFs contribute unequally to task risk (e.g., certain axes require tighter precision or have smaller tolerance margins).
Uniform DoF averaging can therefore be suboptimal.
We introduce nonnegative DoF weights $\beta\in\mathbb{R}^{7}$ to adaptively prioritize kinematically critical axes.
Specifically, we replace uniform DoF aggregation with a weighted step score $\tilde e_t^{(\beta)} = \sum_{d=1}^{7}\beta_d \tilde H_{t,d}$.
The final uncertainty score is then computed using max-based sliding window pooling:
\begin{equation}
S_{\text{SW+ATR+BO}}(\tau) \;=\; \max_{t}\; \frac{1}{w}\sum_{k=t}^{t+w-1}\tilde e_k^{(\beta)}.
\label{eq:swatr_bo_score}
\end{equation}

\paragraph{Bayesian Optimization objective.}
We tune $(w,\alpha,\beta)$ on a calibration split by directly maximizing the discriminative power of the uncertainty score.
Concretely, each rollout $\tau_i$ yields a scalar score $S(\tau_i)$ and a binary label $y_i$; we optimize
\begin{equation}
\label{eq:bo_obj}
(w^\star,\alpha^\star,\beta^\star) = \arg\max_{w,\alpha,\beta}
\mathrm{AUROC}\Big( \{S_{\text{SW+ATR+BO}}(\tau_i)\}, \{1-y_i\} \Big).
\end{equation}
where $\mathrm{AUROC}$ is computed on the calibration set and $(1-y_i)$ treats failures as the positive class.
This objective is expensive to evaluate because it requires evaluating the uncertainty score over a large set of rollouts for each candidate hyperparameter setting.
We therefore use Bayesian Optimization~\cite{snoek2012practical, nogueira_bayesopt}, which models the unknown mapping from hyperparameters to AUROC with a Gaussian process and selects new evaluations by maximizing an acquisition function, enabling sample-efficient search in a continuous parameter space.

\paragraph{Search space.}
We optimize the window length $w\in[10, 100]$, the stability contrast $\alpha\in[0.05,0.95]$, and DoF weights $\beta_d\in[1,10]$ for $d=1,\dots,7$.
In all experiments, the resulting hyperparameters are fixed and reused for evaluation on held-out tasks.

\paragraph{From scores to decisions (thresholding).}
For deployment, $S(\tau)$ is converted into a binary prediction indicating \emph{success} or \emph{failure}.
We select a threshold $\gamma^\star$ on the calibration split and then keep it fixed for testing.
In our experiments, we use Youden's index~\cite{youden1950index} on the ROC curve to choose $\gamma^\star$ (Sec.~\ref{sec:exp:detail}).
We describe this threshold selection in the experimental protocol for clarity, but note that it is compatible with any operating-point selection strategy.

\section{Experiments}
\label{sec:exp}

\subsection{Experiment Details}
\label{sec:exp:detail}

\paragraph{LIBERO task suites.}
We evaluate rollout-level uncertainty quantification on four LIBERO suites:
LIBERO-SPATIAL, LIBERO-OBJECT, and LIBERO-GOAL, which each contain 10 curated tasks designed to isolate transfer of spatial relationships, object identities, and task goals.
In addition, we evaluate on LIBERO-10, a subset of 10 long-horizon tasks drawn from LIBERO-100 (the suite of 100 entangled tasks).
All rollouts are labeled by the simulator as \emph{success} or \emph{failure}.

\paragraph{OpenVLA model.}
We use OpenVLA~\cite{kim2024openvla} as the underlying VLA model.
OpenVLA is a 7B-parameter open-source vision-language-action model trained on a large-scale collection of real-world robot demonstrations and built on a Llama-2 language backbone with a visual encoder that fuses DINOv2 and SigLIP features.
We treat OpenVLA as a black-box model and only use its per-step token distributions to compute entropy-based uncertainty signals (Sec.~\ref{sec:method}).

\paragraph{Train/Test split.}
For each suite, we split rollouts into two disjoint halves:
the train split is used to tune hyperparameters and to select the decision threshold, and the test split is used for reporting.

\paragraph{Metrics.}
We follow common practices in Uncertainty Quantification (UQ)~\cite{hendrycks2016baseline} by treating \emph{failure} as the positive class and reporting:
(i) AUROC on the test split, and
(ii) Accuracy on the test split after thresholding scores using a threshold $\gamma^\star$ chosen on the train split by Youden's index~\cite{youden1950index}:
\begin{equation}
\label{eq:youden_exp}
\begin{aligned}
J(\gamma) \;&=\; \mathrm{TPR}(\gamma)-\mathrm{FPR}(\gamma), \\
\gamma^\star \;&=\; \arg\max_{\gamma} J(\gamma).
\end{aligned}
\end{equation}
Accuracy is computed using $\gamma^\star$ fixed from the train split; thus AUROC reflects separability, while Accuracy reflects a deployment-style discrete trigger.

\paragraph{Computational Overhead.}
The online inference cost of our method is negligible ($\mathcal{O}(1)$ per step). The Sliding Window is implemented as an efficient rolling queue, and Action Transfer Reweighting requires only a lightweight sign comparison. As a result, our framework runs seamlessly at OpenVLA's default control frequency without introducing measurable latency. For the offline BO stage, the procedure is highly sample-efficient: it uses only a small calibration set (e.g., 50 rollouts) and typically converges in fewer than 50 iterations, requiring under 5 minutes on a single RTX 4090 GPU. This modest calibration overhead makes the overall approach highly practical for rapid adaptation to new tasks and environments.

\subsection{Methods Compared}
\label{sec:exp:methods}
We compare our framework against standard uncertainty baselines and progressive ablations.

\paragraph{Global Averaging (Baseline).}
This method follows the standard practice in VLA uncertainty quantification by computing the mean of token-level entropy across all timesteps and degrees of freedom (DoF). As shown in our experiments, this approach suffers from the \textit{Averaging Trap}, where long horizons of successful execution mask localized uncertainty spikes associated with failure.

\paragraph{Sliding Window Pooling (SW).}
SW replaces global averaging with max-based temporal sliding window pooling. By aggregating uncertainty over short windows and retaining the maximum response, SW captures transient high-risk segments that would otherwise be masked by global statistics.

\paragraph{Action Transfer Reweighting (ATR).}
ATR reweights token-level entropy based on action stability, assigning higher weight to timesteps exhibiting oscillatory behavior. This emphasizes uncertainty that coincides with unstable physical execution while down-weighting entropy under smooth motion.

\paragraph{SW+ATR (Combined).}
This variant applies sliding window pooling to stability-weighted entropy, combining temporal locality with motion-aware weighting to capture uncertainty that is both temporally localized and physically relevant.

\paragraph{SW+ATR+BO (DoF-Adaptive Calibration with Bayesian Optimization).}
This method jointly calibrates the sliding window size, stability weighting, and DoF importance by learning task-specific parameters via Bayesian Optimization. Rather than assuming uniform contribution from all joints or fixed temporal and stability settings, it captures a consistent pattern in which certain axes and time scales are more critical for uncertainty prediction, producing a calibrated uncertainty score aligned with the task’s kinematic structure.

\subsection{Main Results}
\label{sec:exp:main}
We report the test AUROC and Accuracy in Table~\ref{tab:main_results}. Our proposed framework consistently outperforms the global averaging baseline across all task suites, indicating that local and adaptive uncertainty estimation plays an important role in robotic introspection.

\paragraph{Performance Overview.} The Baseline method ($\text{mean}_{t,d}$) performs poorly, particularly on long-horizon tasks such as LIBERO-10, where it achieves an AUROC of only 0.468. This empirically confirms the \textit{Averaging Trap}: in long trajectories, the large number of low-entropy tokens obscures the small number of high-entropy tokens that are associated with failure. To ensure a rigorous evaluation, we additionally investigated statistical extrema aggregators (e.g., Global Max, Top-5\% CVaR, Log-Sum-Exp). However, empirical results showed that these extrema methods consistently underperformed the global mean. Successful trajectories naturally exhibit brief, benign entropy spikes during active manipulation, which extrema-based methods fail to isolate from actual failures. Therefore, we report the global mean as our strongest standard baseline. In contrast, our fully calibrated method (SW+ATR+BO) achieves the highest AUROC in 3 out of 4 suites, showing that focused uncertainty extraction is more effective than global aggregation.

\paragraph{Task-Specific Analysis.} On LIBERO-SPATIAL, which is designed to test transfer of spatial relationships with visually similar objects, our method improves AUROC from 0.845 to 0.936. This suggests that uncertainty spikes tend to concentrate around short decision points where the model needs to resolve the correct spatial configuration, which can be masked by global averaging. For LIBERO-OBJECT, which requires identifying and manipulating a specific object type, the SW component alone yields a substantial improvement (e.g., $+0.223$ AUROC on OBJECT). This is consistent with the idea that uncertainty is localized to brief interaction phases, such as approaching, grasping, or lifting, even when the rest of the rollout is stable. For LIBERO-GOAL, where objects and spatial layouts are fixed but task goals differ, the gains from SW suggest that uncertainty is often tied to selecting the correct action sequence to satisfy the goal. Notably, on the challenging LIBERO-10 suite, the addition of BO leads to a clear improvement in performance (0.738 to 0.838). This indicates that for long-horizon, multi-stage tasks, emphasizing the DoFs that are more critical at interaction-heavy phases can improve failure separation.

\begin{table*}[t]
\centering
\caption{Rollout-level failure prediction on LIBERO suites with OpenVLA.
We report test AUROC and test Accuracy (threshold chosen by Youden's index on the train split).
\textbf{Bold} denotes the best result and \underline{underline} denotes the second-best result for each metric within a suite.}
\label{tab:main_results}
\setlength{\tabcolsep}{4.2pt}
\resizebox{\textwidth}{!}{%
\begin{tabular}{lcccccccc}
\toprule
& \multicolumn{2}{c}{LIBERO-SPATIAL} & \multicolumn{2}{c}{LIBERO-OBJECT} & \multicolumn{2}{c}{LIBERO-GOAL} & \multicolumn{2}{c}{LIBERO-10} \\
\cmidrule(lr){2-3}\cmidrule(lr){4-5}\cmidrule(lr){6-7}\cmidrule(lr){8-9}
Method (best params)
& AUROC$\uparrow$ & Acc$\uparrow$
& AUROC$\uparrow$ & Acc$\uparrow$
& AUROC$\uparrow$ & Acc$\uparrow$
& AUROC$\uparrow$ & Acc$\uparrow$ \\
\midrule
Baseline ($\text{mean}_{t,d}$)
& 0.845 & 0.892
& 0.542 & 0.744
& 0.641 & 0.755
& 0.468 & 0.622 \\

SW ($w{=}70$)
& 0.908 & 0.912
& 0.765 & 0.744
& \underline{0.837} & \textbf{0.846}
& 0.699 & 0.691 \\

ATR ($\alpha{=}0.9$)
& 0.847 & 0.868
& 0.567 & \textbf{0.768}
& 0.670 & 0.819
& 0.508 & 0.606 \\

SW+ATR ($w{=}60,\alpha{=}0.9$)
& \underline{0.919} & \textbf{0.948}
& \underline{0.772} & \textbf{0.768}
& \textbf{0.841} & \underline{0.843}
& \underline{0.738} & \underline{0.711} \\
\midrule
SW+ATR+BO ($w^\star,\alpha^\star,\beta^\star$)
& \textbf{0.936} & \underline{0.939}
& \textbf{0.793} & 0.692
& 0.811 & 0.807
& \textbf{0.838} & \textbf{0.743} \\
\bottomrule
\end{tabular}}
\end{table*}

\subsection{Ablation Study}
\label{sec:exp:ablation}

\begin{figure}[t]
    \centering
    \includegraphics[width=\columnwidth]{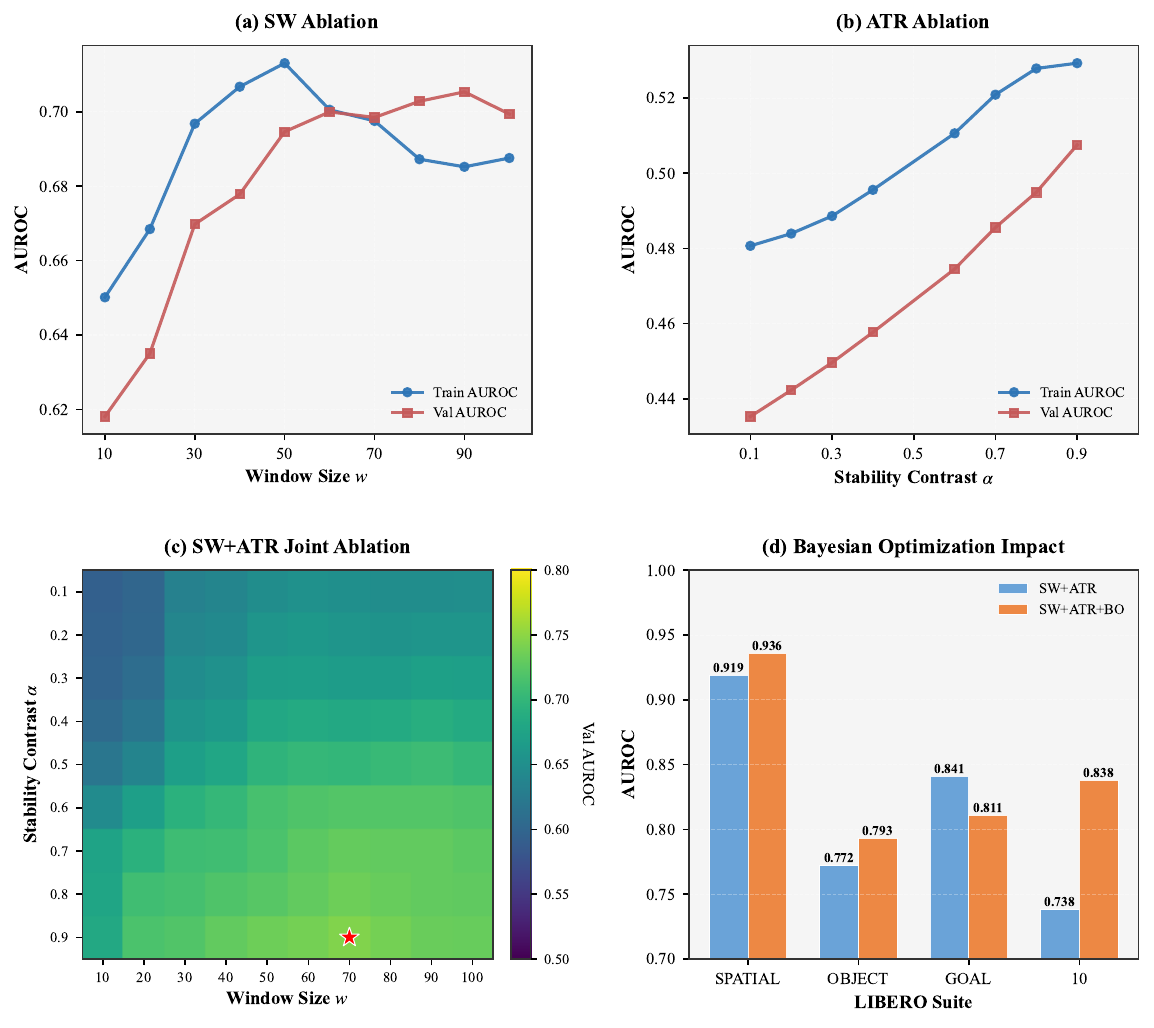}
    \caption{Ablation studies on LIBERO-10.
    (a) SW window size ablation on LIBERO-10.
    (b) ATR stability contrast $\alpha$ ablation on LIBERO-10.
    (c) Joint SW+ATR ablation on LIBERO-10; the red star indicates the best-performing $(w,\alpha)$ pair.
    (d) Effect of DoF-adaptive calibration, comparing SW+ATR with and without Bayesian Optimization across LIBERO suites.}
    \label{fig:ablation_libero10}
\end{figure}

To keep the main paper concise, we present hyperparameter ablations on LIBERO-10 and defer ablations on the other suites (as well as additional thresholded metrics such as Precision/Recall/F1) to the appendix.

\paragraph{SW window size ablation.}
We sweep the sliding-window size $w$ (stride fixed) and report test AUROC in Fig.~\ref{fig:ablation_libero10}(a). Performance improves rapidly as $w$ increases from 10 to 50 and plateaus between $w=50$ and $w=90$. This suggests that very short windows are sensitive to benign noise, while a window of approximately 50--90 steps (roughly 2.5--4.5 seconds) is optimal for capturing the duration of atomic actions in failure sequences.

\paragraph{ATR weight ablation.}
We sweep the ATR contrast parameter $\alpha$ and report test AUROC in Fig.~\ref{fig:ablation_libero10}(b). We observe a monotonic increase in performance as $\alpha$ grows, with the highest accuracy achieved at $\alpha=0.9$. This confirms that high-frequency action oscillation is a strong indicator of failure, and heavily weighting these unstable moments significantly helps to distinguish uncertainty.

\paragraph{SW+ATR joint ablation.}
We evaluate a grid over $(w,\alpha)$ and visualize test AUROC as a heatmap in Fig.~\ref{fig:ablation_libero10}(c). The results show a synergistic effect between the two components. The global maximum (marked by the red star) is located at $w=70$ and $\alpha=0.9$, indicating that the model performs best when it combines a sufficiently large temporal receptive field with a high sensitivity to physical instability.

\subsection{Bayesian-Optimized Hyperparameters}
\label{sec:exp:bo_hparams}

The hyperparameters learned via Bayesian Optimization provide information about the physical sources of task risk (we detail the optimal parameters in Appendix, see Table~\ref{tab:bo_hparams_app}). Visualizing the learned DoF weights $\beta^\star$ (Appendix Fig.~\ref{fig:dof_weights_app}) makes cross-suite patterns more apparent. We observe a consistent pattern in which the gripper dimension, and in several suites the vertical axis ($\Delta z$), receive the highest weights. This is consistent with the observation that manipulation failures often occur during object interaction phases, such as missing a grasp or colliding during lifting. Additionally, for LIBERO-OBJECT and LIBERO-GOAL, the model assigns the highest weight to the pitch rotation ($\Delta \text{pitch}$). In LIBERO-OBJECT, this suggests that orientation-related uncertainty is particularly informative when transferring to novel object geometries, while in LIBERO-GOAL it reflects the role of precise end-effector orientation in satisfying goal-specific motion and placement constraints. These results indicate that the method assigns higher importance to kinematic axes that are more informative for failure prediction.

\subsection{Zero-Shot Generalization Across Suites}
\label{sec:exp:zeroshot}

To address potential concerns regarding overfitting during Bayesian Optimization tuning, we conducted a comprehensive cross-suite transfer experiment. We evaluated the optimal parameters $(w^\star, \alpha^\star, \beta^\star)$ calibrated exclusively on the LIBERO-SPATIAL train split directly on the held-out test splits of the remaining three suites in a zero-shot manner.

\begin{table}[h]
\centering
\caption{Cross-suite zero-shot transfer test AUROC. Parameters calibrated solely on the LIBERO-SPATIAL train split transfer robustly to unseen suites.}
\label{tab:cross_suite}
\vspace{0.3em}
\setlength{\tabcolsep}{8pt}
\begin{tabular}{lccc}
\toprule
Target Suite & Global Mean & \textbf{Zero-Shot Transfer} & Oracle BO \\
\midrule
LIBERO-OBJECT & 0.542 & \textbf{0.780} & 0.793 \\
LIBERO-GOAL   & 0.641 & \textbf{0.819} & 0.811 \\
LIBERO-10     & 0.468 & \textbf{0.715} & 0.838 \\
\bottomrule
\end{tabular}
\end{table}

As shown in Table~\ref{tab:cross_suite}, the zero-shot transfer consistently outperforms the global mean baseline across all unseen suites. Notably, on LIBERO-OBJECT and LIBERO-GOAL, the transferred parameters perform on par with or even slightly exceed the suite-specific Oracle BO. This confirms that our framework learns generalizable physical uncertainty principles, including the prioritization of gripper dimensions and the penalization of high-frequency oscillations, rather than overfitting to specific tasks. This robust cross-suite transferability underscores the real-world deployment potential of our approach.

\section{Conclusion}
\label{sec:conclusion}

In this work, we identified the \textit{Averaging Trap} as a fundamental limitation in applying standard LLM uncertainty metrics to continuous robotic control, where transient failure signals are often masked by global statistics. To address this, we proposed a unified uncertainty quantification framework that shifts the focus from global averaging to local, physically grounded risk detection. By integrating max-based Sliding Window Pooling (SW) to capture temporal spikes, Action Transfer Reweighting (ATR) to penalize oscillatory behavior, and Bayesian Optimization (BO) for adaptive DoF calibration, our method effectively isolates critical moments of instability.

Evaluations on the LIBERO benchmark demonstrate that our approach significantly outperforms standard entropy baselines in distinguishing successful rollouts from failures. These improvements provide a more reliable signal for human-in-the-loop intervention, a prerequisite for deploying Vision-Language-Action models in real-world environments. Future work will focus on leveraging these calibrated uncertainty signals not just for passive detection, but for active uncertainty-aware planning and closed-loop error recovery.

\bibliographystyle{alpha}
\bibliography{main}

\clearpage
\appendix
\section{Additional Experimental Details}
\label{sec:appendix}
\setcounter{table}{0}
\renewcommand{\thetable}{A\arabic{table}}
\setcounter{figure}{0}
\renewcommand{\thefigure}{A\arabic{figure}}

\subsection{Bayesian-Optimized Hyperparameters Details}
\label{sec:appendix:bo_hparams}

This section provides the detailed hyperparameters $(w^\star, \alpha^\star, \beta^\star)$ optimized via Bayesian Optimization on the train split of each suite, as discussed in Section~\ref{sec:exp:bo_hparams}. We also provide a visualization of the learned Degrees of Freedom (DoF) weights $\beta^\star$.

\begin{table}[H]
\centering
\caption{Best SW+ATR+BO hyperparameters found on the train split of each suite.
All $\beta^\star$ values are rounded to one decimal place for readability.}
\label{tab:bo_hparams_app}
\renewcommand{\arraystretch}{1.2}
\setlength{\tabcolsep}{7pt}
\begin{tabular}{lccc}
\toprule
Suite & $w^\star$ & $\alpha^\star$ & $\boldsymbol{\beta}^\star$ (7-DoF) \\
\midrule
LIBERO-SPATIAL & 93  & 0.95 & [1.0, 2.6, 10.0, 1.0, 1.0, 1.0, 10.0] \\
LIBERO-OBJECT  & 62  & 0.05 & [2.6, 1.0, 1.2, 1.0, 10.0, 7.5, 10.0] \\
LIBERO-GOAL    & 72  & 0.05 & [1.0, 1.0, 4.6, 1.0, 10.0, 1.0, 10.0] \\
LIBERO-10      & 100 & 0.95 & [1.6, 6.9, 10.0, 1.0, 1.0, 1.0, 10.0] \\
\bottomrule
\end{tabular}
\end{table}

\begin{figure}[H]
  \centering
  \includegraphics[width=0.75\textwidth]{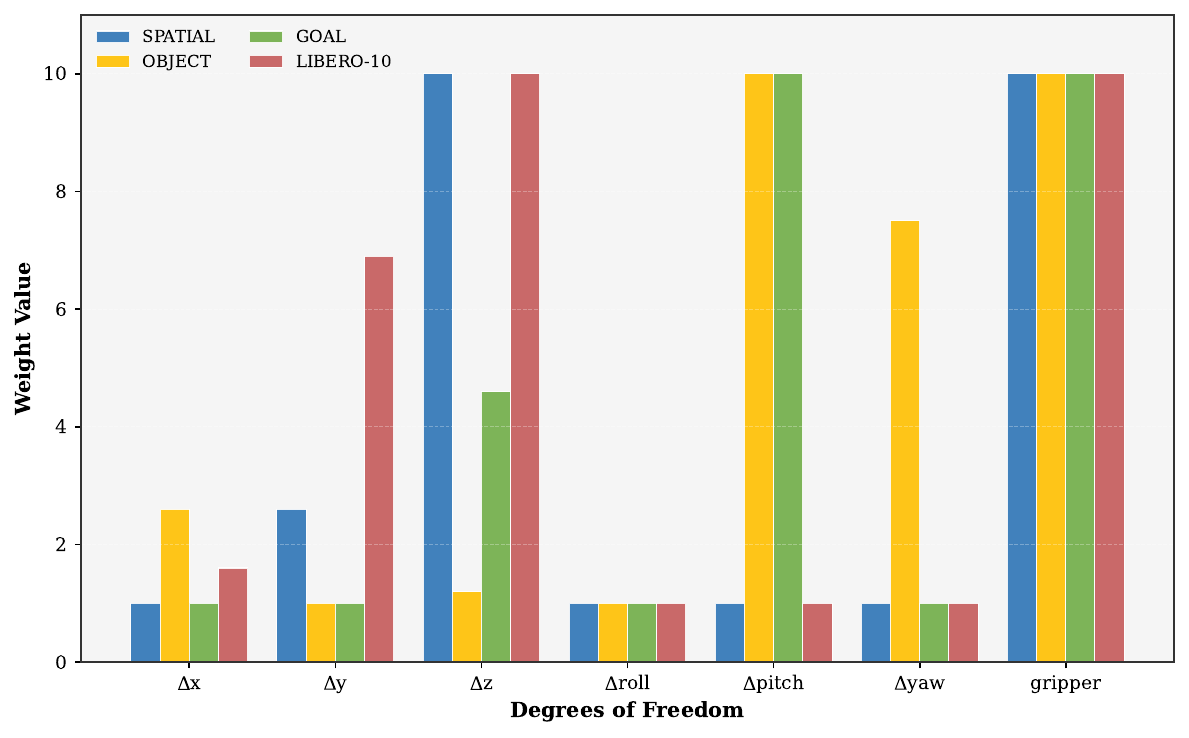}
  \caption{DoF importance learned by Bayesian Optimization.
  We visualize the optimized DoF weights $\beta^\star$ across LIBERO suites.
  Gripper and $\Delta z$ consistently receive high weights, while $\Delta\text{pitch}$ is emphasized in LIBERO-OBJECT and LIBERO-GOAL, supporting DoF-adaptive calibration.}
  \label{fig:dof_weights_app}
\end{figure}

\subsection{Benchmark Success Rates per Task}
\label{sec:appendix:success_rates}

This section reports the raw task-level success rates of the OpenVLA model on the LIBERO benchmark.
Each suite contains 10 tasks with 50 evaluation episodes per task.
These statistics provide additional context for the uncertainty quantification results reported in the main paper, and help explain task difficulty variations across suites.

\begin{figure}[H]
    \centering
    \includegraphics[width=0.75\textwidth]{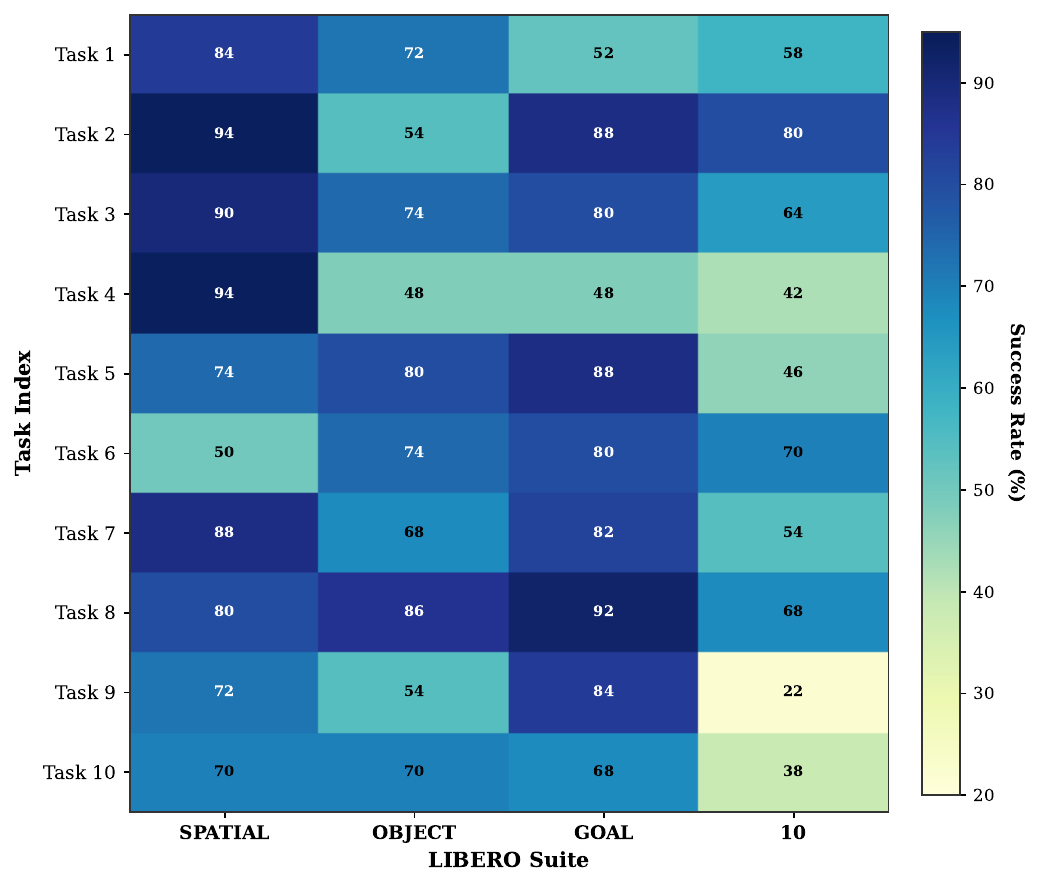}
    \caption{Per-task success rates (\%) of OpenVLA baseline across LIBERO suites.
    Darker cells indicate higher success. Task difficulty varies significantly both across and within suites, motivating the need for reliable failure prediction via uncertainty quantification.}
    \label{fig:appendix:success_heatmap}
\end{figure}

We visualize the success rates in Fig.~\ref{fig:appendix:success_heatmap}. SPATIAL and GOAL suites generally exhibit higher success rates, while LIBERO-10 remains the most challenging, with frequent long-horizon or multi-stage failures. This variation further motivates our focus on task-level uncertainty estimation and robust failure prediction. The detailed per-task results are listed below for each suite.

\subsubsection{LIBERO-SPATIAL}

\noindent\textbf{Overall Success Rate:} 79.6\% (398/500)

\vspace{0.3em}
\begin{tabularx}{\linewidth}{@{}Xr@{}}
Task 1: pick up the black bowl between plate \& ramekin and place it on the plate & 84.0\% \\
Task 2: pick up the black bowl next to the ramekin and place it on the plate & 94.0\% \\
Task 3: pick up the black bowl from table center and place it on the plate & 90.0\% \\
Task 4: pick up the black bowl on the cookie box and place it on the plate & 94.0\% \\
Task 5: pick up the black bowl in the top drawer of the wooden cabinet and place it on the plate & 74.0\% \\
Task 6: pick up the black bowl on the ramekin and place it on the plate & 50.0\% \\
Task 7: pick up the black bowl next to the cookie box and place it on the plate & 88.0\% \\
Task 8: pick up the black bowl on the stove and place it on the plate & 80.0\% \\
Task 9: pick up the black bowl next to the plate and place it on the plate & 72.0\% \\
Task 10: pick up the black bowl on the wooden cabinet and place it on the plate & 70.0\% \\
\end{tabularx}

\vspace{1em}

\subsubsection{LIBERO-OBJECT}

\noindent\textbf{Overall Success Rate:} 68.0\% (340/500)

\vspace{0.3em}
\begin{tabularx}{\linewidth}{@{}Xr@{}}
Task 1: pick up the alphabet soup and place it in the basket & 72.0\% \\
Task 2: pick up the cream cheese and place it in the basket & 54.0\% \\
Task 3: pick up the salad dressing and place it in the basket & 74.0\% \\
Task 4: pick up the bbq sauce and place it in the basket & 48.0\% \\
Task 5: pick up the ketchup and place it in the basket & 80.0\% \\
Task 6: pick up the tomato sauce and place it in the basket & 74.0\% \\
Task 7: pick up the butter and place it in the basket & 68.0\% \\
Task 8: pick up the milk and place it in the basket & 86.0\% \\
Task 9: pick up the chocolate pudding and place it in the basket & 54.0\% \\
Task 10: pick up the orange juice and place it in the basket & 70.0\% \\
\end{tabularx}

\vspace{1em}

\subsubsection{LIBERO-GOAL}

\noindent\textbf{Overall Success Rate:} 76.2\% (381/500)

\vspace{0.3em}
\begin{tabularx}{\linewidth}{@{}Xr@{}}
Task 1: open the middle drawer of the cabinet & 52.0\% \\
Task 2: put the bowl on the stove & 88.0\% \\
Task 3: put the wine bottle on top of the cabinet & 80.0\% \\
Task 4: open the top drawer and put the bowl inside & 48.0\% \\
Task 5: put the bowl on top of the cabinet & 88.0\% \\
Task 6: push the plate to the front of the stove & 80.0\% \\
Task 7: put the cream cheese in the bowl & 82.0\% \\
Task 8: turn on the stove & 92.0\% \\
Task 9: put the bowl on the plate & 84.0\% \\
Task 10: put the wine bottle on the rack & 68.0\% \\
\end{tabularx}

\vspace{1em}

\subsubsection{LIBERO-10}

\noindent\textbf{Overall Success Rate:} 54.2\% (271/500)

\vspace{0.3em}
\begin{tabularx}{\linewidth}{@{}Xr@{}}
Task 1: put both the alphabet soup and the tomato sauce in the basket & 58.0\% \\
Task 2: put both the cream cheese box and the butter in the basket & 80.0\% \\
Task 3: turn on the stove and put the moka pot on it & 64.0\% \\
Task 4: put the black bowl in the bottom drawer of the cabinet and close it & 42.0\% \\
Task 5: place mugs on left/right plates & 46.0\% \\
Task 6: pick up the book and place it in the caddy & 70.0\% \\
Task 7: place mug and pudding beside plate & 54.0\% \\
Task 8: put both soup and cream cheese in the basket & 68.0\% \\
Task 9: put both moka pots on the stove & 22.0\% \\
Task 10: microwave the yellow and white mug and close it & 38.0\% \\
\end{tabularx}

\subsection{Detailed Ablation and Metric Breakdown}
\label{sec:appendix:ablation}

This appendix provides additional experimental results referenced in the main paper.
Specifically, we present detailed hyperparameter ablation studies and metric breakdowns
(AUROC, Accuracy, Precision, Recall, and F1-score) for the
LIBERO-SPATIAL, LIBERO-OBJECT, LIBERO-GOAL, and
LIBERO-10 task suites.

\begin{figure}[h]
    \centering
    \includegraphics[width=\textwidth]{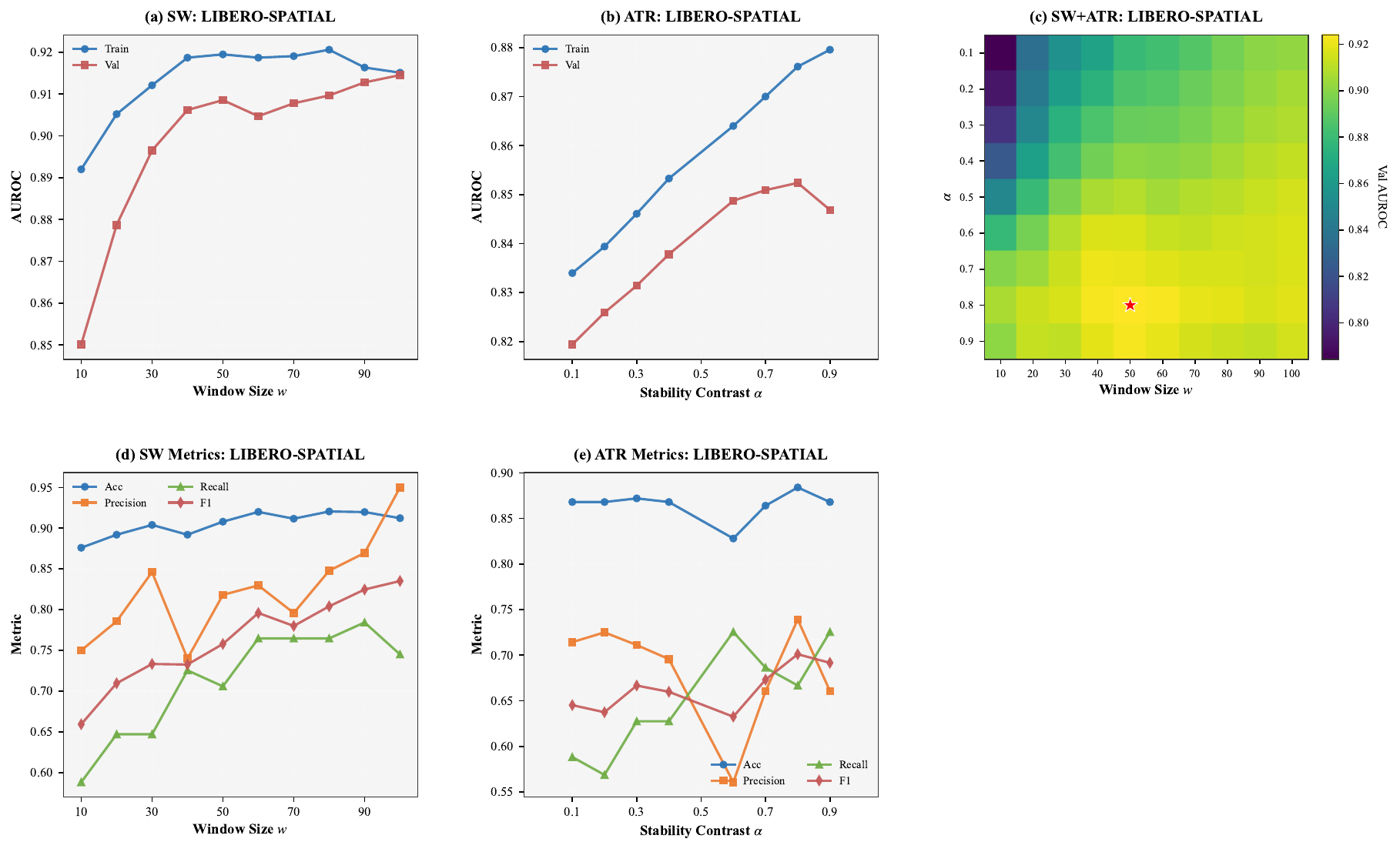}
    \caption{Ablation Study on LIBERO-SPATIAL. 
    (a) Impact of window size $w$ on AUROC. 
    (b) Impact of stability contrast $\alpha$ on AUROC. 
    (c) Joint heatmap of AUROC across $w$ and $\alpha$. 
    (d-e) Detailed metrics (Accuracy, Precision, Recall, F1) for SW and ATR components respectively.}
    \label{fig:ablation_spatial}
\end{figure}

\begin{figure}[h]
    \centering
    \includegraphics[width=\textwidth]{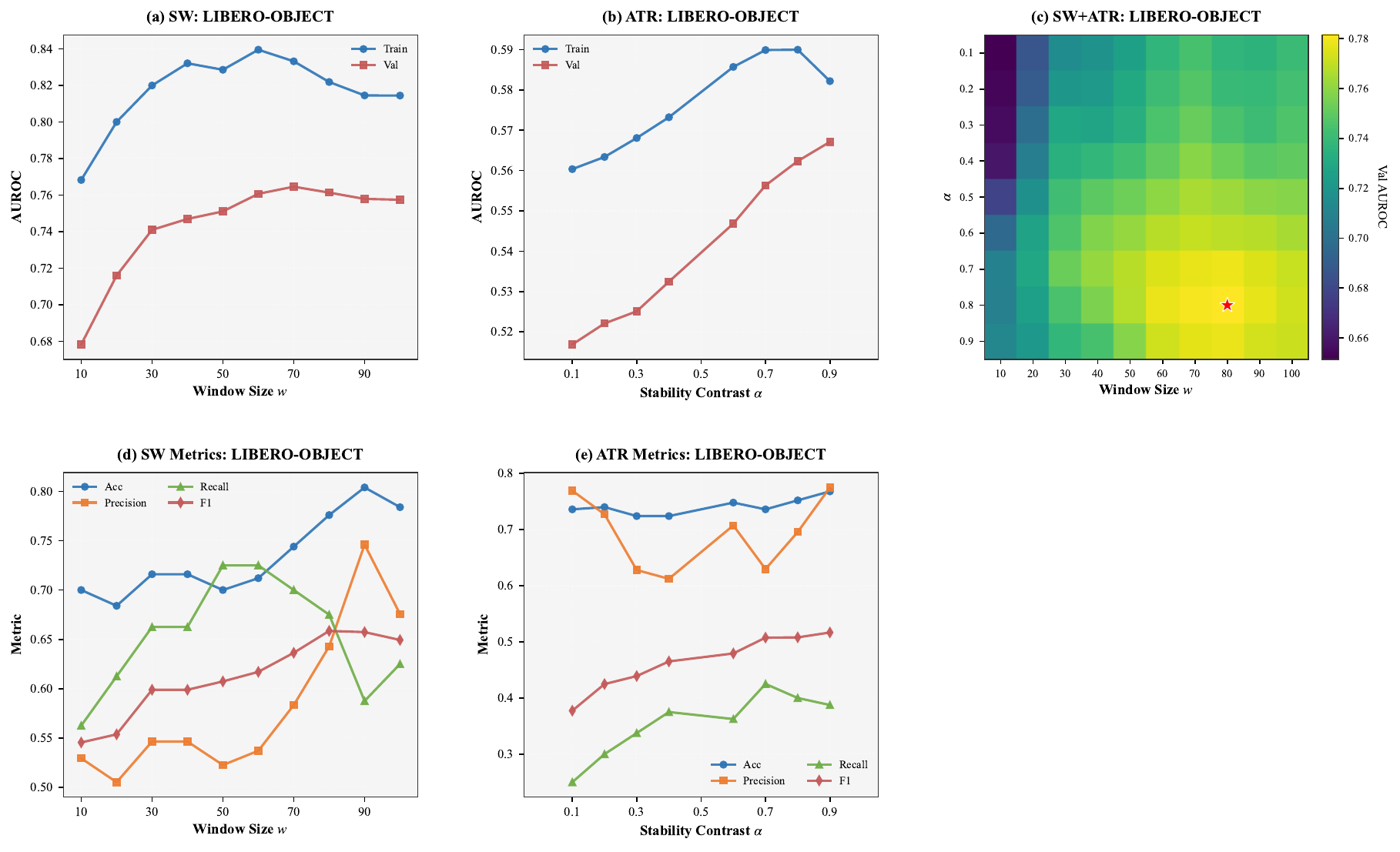}
    \caption{Ablation Study on LIBERO-OBJECT. 
    (a) Impact of window size $w$ on AUROC. 
    (b) Impact of stability contrast $\alpha$ on AUROC. 
    (c) Joint heatmap of AUROC across $w$ and $\alpha$. 
    (d-e) Detailed metrics (Accuracy, Precision, Recall, F1) for SW and ATR components respectively.}
    \label{fig:ablation_object}
\end{figure}

\begin{figure}[h]
    \centering
    \includegraphics[width=\textwidth]{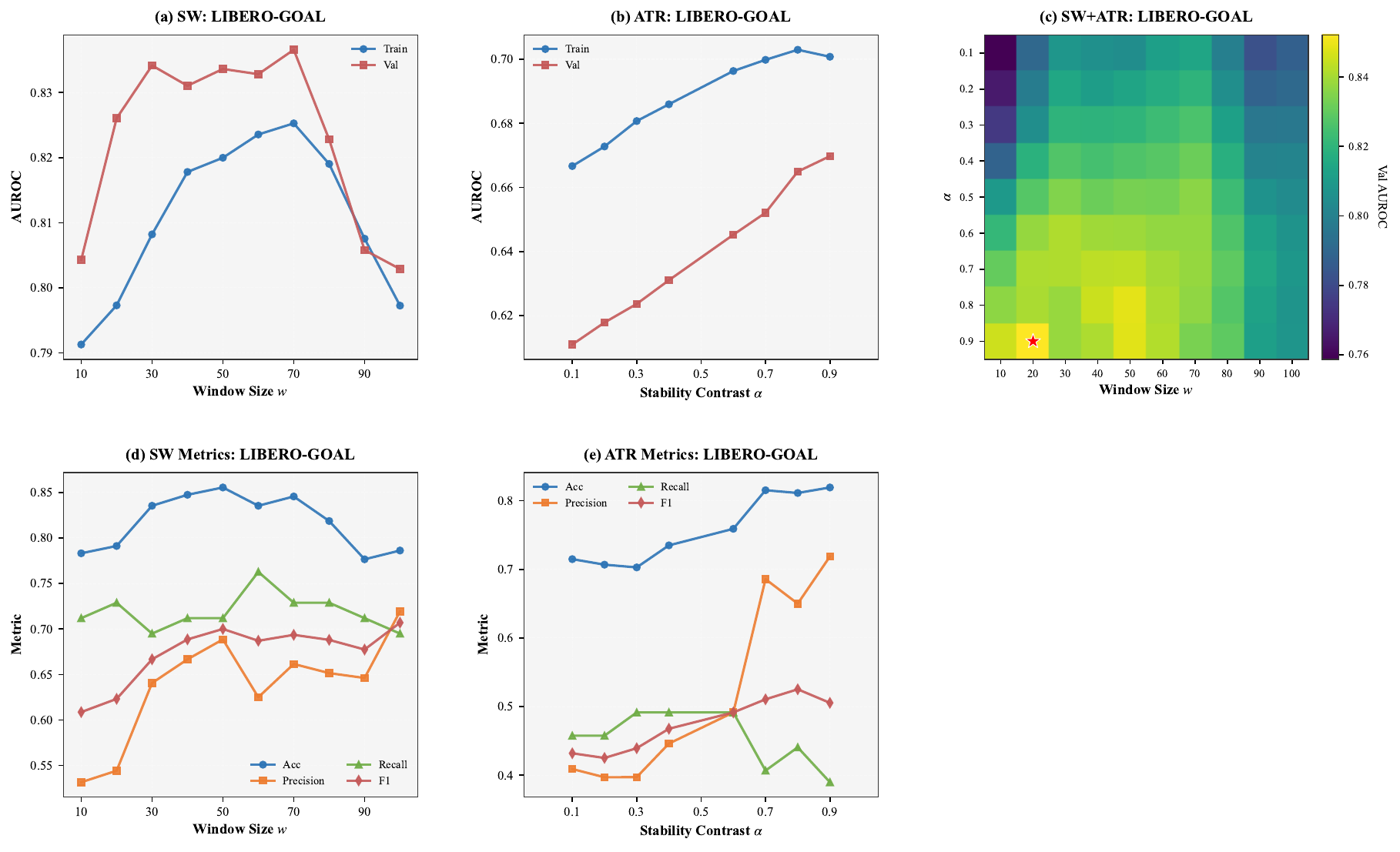}
    \caption{Ablation Study on LIBERO-GOAL. 
    (a) Impact of window size $w$ on AUROC. 
    (b) Impact of stability contrast $\alpha$ on AUROC. 
    (c) Joint heatmap of AUROC across $w$ and $\alpha$. 
    (d-e) Detailed metrics (Accuracy, Precision, Recall, F1) for SW and ATR components respectively.}
    \label{fig:ablation_goal}
\end{figure}

\begin{figure}[h]
    \centering
    \includegraphics[width=\textwidth]{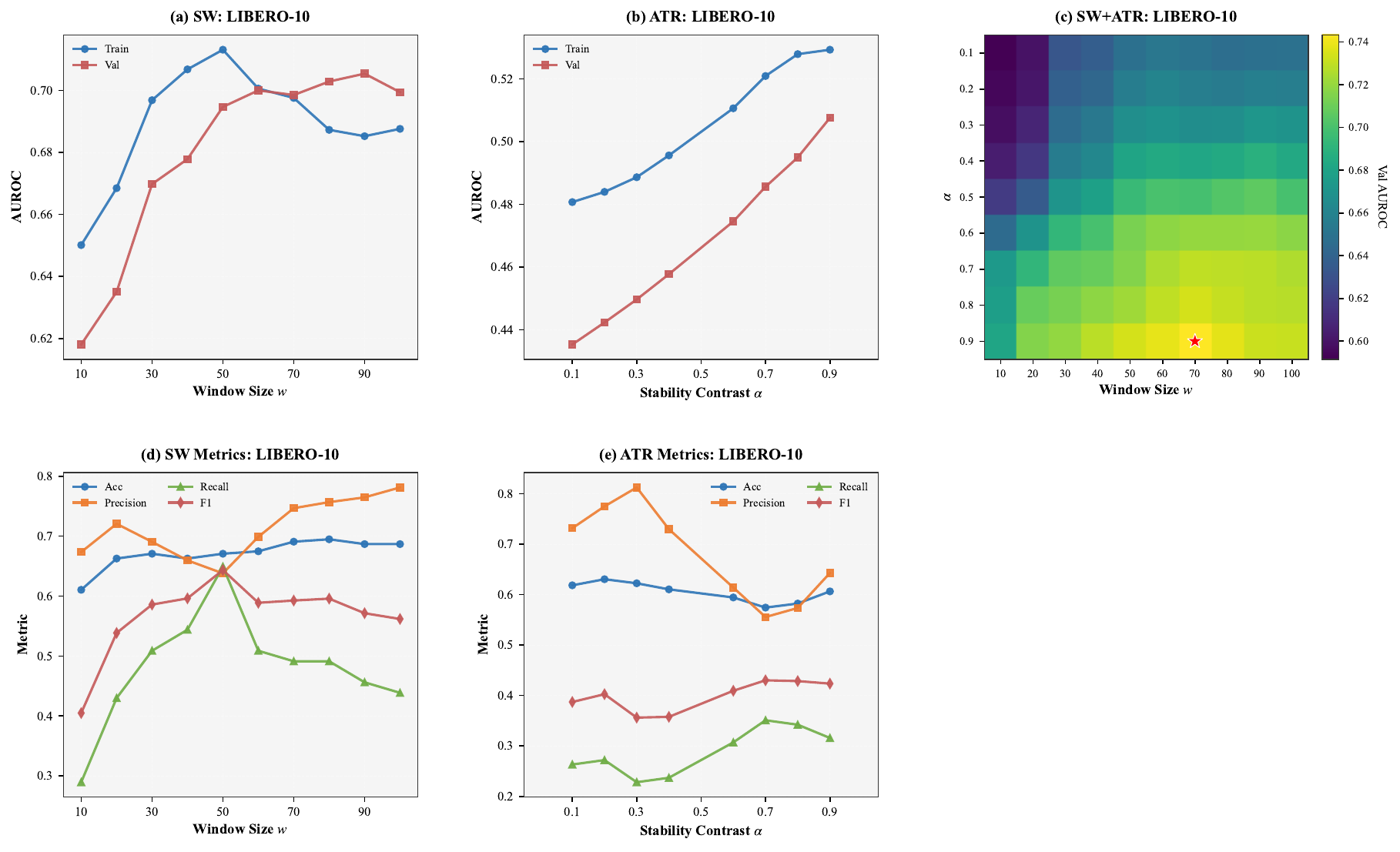}
    \caption{Ablation Study on LIBERO-10. 
    (a) Impact of window size $w$ on AUROC. 
    (b) Impact of stability contrast $\alpha$ on AUROC. 
    (c) Joint heatmap of AUROC across $w$ and $\alpha$. 
    (d-e) Detailed metrics (Accuracy, Precision, Recall, F1) for SW and ATR components respectively.}
    \label{fig:ablation_10}
\end{figure}

\end{document}